\documentclass[10pt,journal,compsoc]{IEEEtran}

\usepackage[utf8]{inputenc} 
\usepackage[T1]{fontenc}  
\usepackage{hyperref}    
\usepackage{url}      
\usepackage{booktabs}    
\usepackage{amsfonts}    
\usepackage{nicefrac}    
\usepackage{microtype}   
\usepackage{lipsum}
\usepackage{graphicx} 
\usepackage{float} 

\usepackage{amsthm,amsmath,amssymb,mathrsfs}
\usepackage{makecell}
\usepackage{multirow}
\usepackage{algorithm,algorithmic}

\begin{document}
\title{Multitask Learning for \\Emotion and Personality Detection}

\author{
 Yang Li,
 Amirmohammad Kazameini,
 Yash Mehta,
 Erik Cambria
 \IEEEcompsocitemizethanks{\IEEEcompsocthanksitem Yang Li is with Northwestern Polytechnical University, China, Email: yang.li@ntu.edu.sg
 \IEEEcompsocthanksitem Amirmohammad Kazameini is with Western University, Canada, Email: akazemei@uwo.ca
 \IEEEcompsocthanksitem Yash Mehta is with the University College London, UK, Email: y.mehta@ucl.ac.uk
 \IEEEcompsocthanksitem Erik Cambria is with Nanyang Technological University, Singapore, Email: cambria@ntu.edu.sg}
\thanks{Corresponding author: Erik Cambria}
}
 
 \markboth{IEEE Transaction on Affective Computing,~Vol.~1, No.~1,~January~2021}{Shell
\MakeLowercase{\textit{et al.}}: Personality Detection with Emotion via Multitask learning}

\IEEEtitleabstractindextext{
\begin{abstract}
In recent years, deep learning-based automated personality trait detection has received a lot of attention, especially now, due to the massive digital footprints of an individual. Moreover, many researchers have demonstrated that there is a strong link between personality traits and emotions. In this paper, we build on the known correlation between personality traits and emotional behaviors, and propose a novel multitask learning framework, SoGMTL that simultaneously predicts both of them. We also empirically evaluate and discuss different information-sharing mechanisms between the two tasks. To ensure the high quality of the learning process, we adopt a MAML-like framework for model optimization. Our more computationally efficient CNN-based multitask model achieves the state-of-the-art performance across multiple famous personality and emotion datasets, even outperforming Language Model based models.

\end{abstract}

\begin{IEEEkeywords}
Personality Trait Detection, Multitask Learning, Information Sharing 
\end{IEEEkeywords}}
 
\maketitle
\IEEEdisplaynontitleabstractindextext
\IEEEpeerreviewmaketitle
\IEEEraisesectionheading{\section{Introduction}\label{sec:introduction}}
Personality traits refer to the difference among individuals in characteristic patterns of thinking, feeling, and behaving. Specifically, personality traits have been related to individual well being, social-institutional outcomes (e.g., occupational choices, job success), and interpersonal (e.g., relationship satisfaction). In addition, in modern times, people prefer delivering their thoughts, emotions, and complaints on the social media platform, e.g., Facebook and Twitter~\cite{sushou}. Hence, there is a widespread interest to develop models that can use online data on human preferences and behavior (i.e., digital footprints) to automatically predict individuals' levels of personality traits for use in job screenings~\cite{liem2018psychology}, recommender-systems~\cite{li2019learning} and social network analysis. You et al.~\cite{youyou2015computer} found that the digital footprint on social media can be used to measure personality traits well. There are different systems in the personality trait description, and the most widely used is called the Five-Factor Model~\cite{goldberg1993structure}. This system includes five traits that can be remembered by the acronym OCEAN: Openness (OPN), Conscientiousness (CON), Extraversion (EXT), Agreeableness (AGR), and Neuroticism (NEU). With the advancement in machine learning research and the availability of larger amounts of data, the ability and desire to detect user personality and preference are now higher than ever. While the performance of these models is not high enough to allow for the precise distinction of people based on their traits, predictions can still be `right' on average and be utilized for digital mass persuasion~\cite{matz2017psychological}. However, automated personality prediction also raises serious concerns with regard to individual privacy and the conception of informed consent~\cite{matz2020privacy}.

Recent works on this have made significant strides in machine learning-based personality detection~\cite{mehta2020icdm,argamon2005lexical,carducci2018twitpersonality,kalghatgi2015neural,liu2017language,youyou2015computer}. However, all existing works are single-task learning in the supervised learning way. For example, Kalghatgi et al.~\cite{kalghatgi2015neural} add an MLP over handcrafted features to do the detection, and Tandera et al.~\cite{tandera2017personality} combined LSTM and CNN together to make a better feature extraction pipeline, and finally Mehta et al.~\cite{mehta2020icdm} combine language models with psycholinguistic features for personality prediction. 
 
As we know, emotion has a direct link to personality. According to work from~\cite{deyoung2007between,revelle2009personality,komulainen2014effect,hiebler2018personality}, we know that neuroticism predicts higher negative emotion and lower positive affect, and conscientiousness, by contrast, is inversely associate with negative emotions, and agreeableness predicts higher positive emotion and lower negative affect, and extraversion is associated with higher positive affect and more positive subjective evaluations of daily activities, and openness is associate with a mix of positive and negative emotions~\cite{barford2016openness}. Therefore, we build on the results showing that personality trait detection and emotion detection are complementary. After exploring different information-sharing mechanisms (e.g., SiG, SiLG, CAG, and SoG) between personality trait detection and emotion detection, we propose a novel multitask learning framework SoGMTL based on CNN for simultaneously detecting personality traits and emotions, and our model achieves the state of the art across both personality and emotion datasets. 
Also, to ensure the high quality of the learning procedure, we propose a MAML-like method for model optimization. 

The rest of the paper is structured as follows: Section~\ref{sec:related_works} introduces related works about personality trait detection and multitask learning; Section~\ref{sec:model} presents the proposed model and discusses several different information sharing gates; Section~\ref{sec:exp} conducts the experiments on multitask learning; finally, Section~\ref{sec:con} concludes the paper and presents future works.
 
\section{Related Works}
\label{sec:related_works}
This paper mainly focuses on personality trait detection with a multitask learning framework. In the framework, based on the observation that personality and emotion are complementary, we select emotion prediction as to the auxiliary task. Therefore, in the related works, we will only give a comprehensive review of personality trait detection and multitask learning.

\subsection{Personality Trait Detection}
 
It has been confirmed by researches that online behavior is related to personality~\cite{hughes2012tale,gosling2011manifestations}. Many works have successfully applied the machine learning methods to detect the personality trait in the content generated over the social media~\cite{celli2014workshop,tkalvcivc2014preface}. Especially in the work of You et al.~\cite{youyou2015computer}, they found that the analysis based on digital footprint was better at measuring personality traits than close others or acquaintances (friends, family, spouse, colleagues, etc.). Personality trait detection can be based on the different types of features, such as text data (self-description, social media content, etc.), demography data (gender, age, followers, etc.), etc. One of the initial models was by Argamon et al.~\cite{argamon2005lexical}, which applied an SVM over the extracted statistical features of functional lexicons to detect the personality trait. Following this work, Farnadi et al.~\cite{farnadi2013well} adopted SVM to make personality detection over the features of network size, density, frequency of updating status, etc. Zhusupova et al.~\cite{jusupova2016characterizing} detected the personality trait of Portuguese users on the Twitter platform based on demographic and social activity information. Kalghatgi et al.~\cite{kalghatgi2015neural} detected the personality trait based on the neural networks (MLP) with the hand-crafted features. Su et al.~\cite{su2016exploiting} applied the RNN and HMM to obtain the personality trait based on the Chinese LIWC annotations extracted from the dialogue. Carducci et al.~\cite{carducci2018twitpersonality} also applied the SVM to do the personality detection, the difference between Farnadi's work~\cite{farnadi2013well} is that the feature they applied is the text data. 

Researchers also leveraged some recent developments of NLP in this field.
Tandera et al.~\cite{tandera2017personality} made personality detection over the text data directly based on deep learning methods (LSTM+CNN). 
At the same time, Liu et al.~\cite{liu2017language} built a hierarchical structure based on Bi-RNN to learn the word and sentence representations that can infer the personality traits from three languages, i.e., English, Italian, and Spanish. Majumder et al.~\cite{majumder2017deep} proposed a CNN-based model to extract fixed-length features from personal documents, and then connected the learned features with 84 additional features in the Mairesse's library for personality feature detection. Van et al.~\cite{van2017personality} tried to infer the personality trait based on the 275 profiles on LinkedIn, a job-related social media platform, and they concluded that extroversion could be well inferred from the self-expression of the profiles. Amirhosseini et al.~\cite{amirhosseini2020machine} increased the accuracy of the MBTI dataset with Gradient boosting. Lynn et al.~\cite{lynn2020hierarchical} used message level attention instead of word-level attention to improving the result by focusing on the most relevant Facebook posts obtained from Kosinski et al dataset~\cite{kosinski2013private}. Gjurkovic et al.~\cite{gjurkovic2020pandora} used Sentence BERT~\cite{reimers2019sentence} to set a benchmark for their huge Reddit dataset named PANDORA, including three personality tests' labels OCEAN, MBTI, and Enneagram. Pandora tried to improve and address the deficiencies of its older version, MBTI19k~\cite{gjurkovic2018reddit}. Kazemeini et al.~\cite{kazameinipersonality} feed BERT embeddings into an SVM based ensemble method to improve the accuracy of Essays dataset of Pennebaker et al.~\cite{pennebaker1999linguistic}. Finally, Mehta et al~\cite{mehta2020icdm} perform a thorough empirical investigation using Language Models (LM) and a variety of psycholinguistic features to identify the best combination and the impact of each feature on predicting the traits, achieving the state-of-the-art results on the Essays as well as the Kaggle MBTI dataset. Recently, Mehta et al.~\cite{mehta2020recent} reviewed the latest advances in deep learning-based automated personality with a focus on effective multimodal personality prediction. 
				
\subsection{Multitask Learning}
Multitask learning has attracted lots of attention in recent years, it is to enhance the performance by learning the commonalities and differences among different tasks~\cite{caruana1997multitask,ruder2017overview,majsen}. 
Typically, one of the widely used models is proposed by Carunana~\cite{caruana1997multitask,caruana93multitasklearning}. And there is a shared-bottom structure across tasks. 
This structure reduces the risk of overfitting, but it will cause optimization conflicts due to task differences.
Recently, some works attempt to resolve such conflicts by adding constraints on specific task parameters~\cite{duong2015low,misra2016cross}.
For example, Duong et al.~\cite{duong2015low} trained two different tasks with different data sets and added L-2 constraints between the two sets of parameters to enhance the adaptability of the model over the low resource data. 
Misra et al.~\cite{misra2016cross} proposed the cross-stitch networks to learn a linear combination of task-specific hidden-layer embeddings. And the tasks on the semantic segmentation and surface normal prediction over the image data validate the effectiveness of this method. 
Yang et al.~\cite{yang2016deep} generated the task-specific parameters by tensor factorization, and it achieved better performance on the task which may lead to conflicts. The drawback of this model relies on its large number of parameters which need more training data. 
Ma et al.~\cite{ma2018modeling} designed a multigate mixture-of-experts based on the work~\cite{jacobs1991adaptive}, it requires several experts (e.g., there are 8 experts in their settings.) to capture different features of the task, then chooses the highest gate score as the final feature. Same as the works from Yang~\cite{yang2016deep}, there is a large number of parameters that need more training data.
Unlike previous works, this paper proposes a task-specific multitask learning framework that is used for personality trait detection and emotion detection.

\section{Model Description }
\label{sec:model}

In this section, we describe our proposed model, with the framework shown in Figure~\ref{fig:structure}. 
Typically, there will be a shared bottom in the structure, and as we have discussed, it is difficult to optimize the model due to different tasks. 
Therefore, based on CNN, this paper designs an efficient information flow pipeline to realize the information exchange between different tasks.

\subsection{Preliminary}
Generally, after the embedding layer, each word in the sentence is represented by the vector $x_{i}\in R^{d}$ where $d$ is the embedding dimension. And the whole sentence is represented by $X=[x_{1},x_{2},...,x_{n}]$ where $n$ is the word number. 
There are three layers in the CNN model, namely the convolutional layer, the pooling layer, and the dense layer.
 
\begin{figure}[htp]
 \centering
   \includegraphics[width=0.9\linewidth]{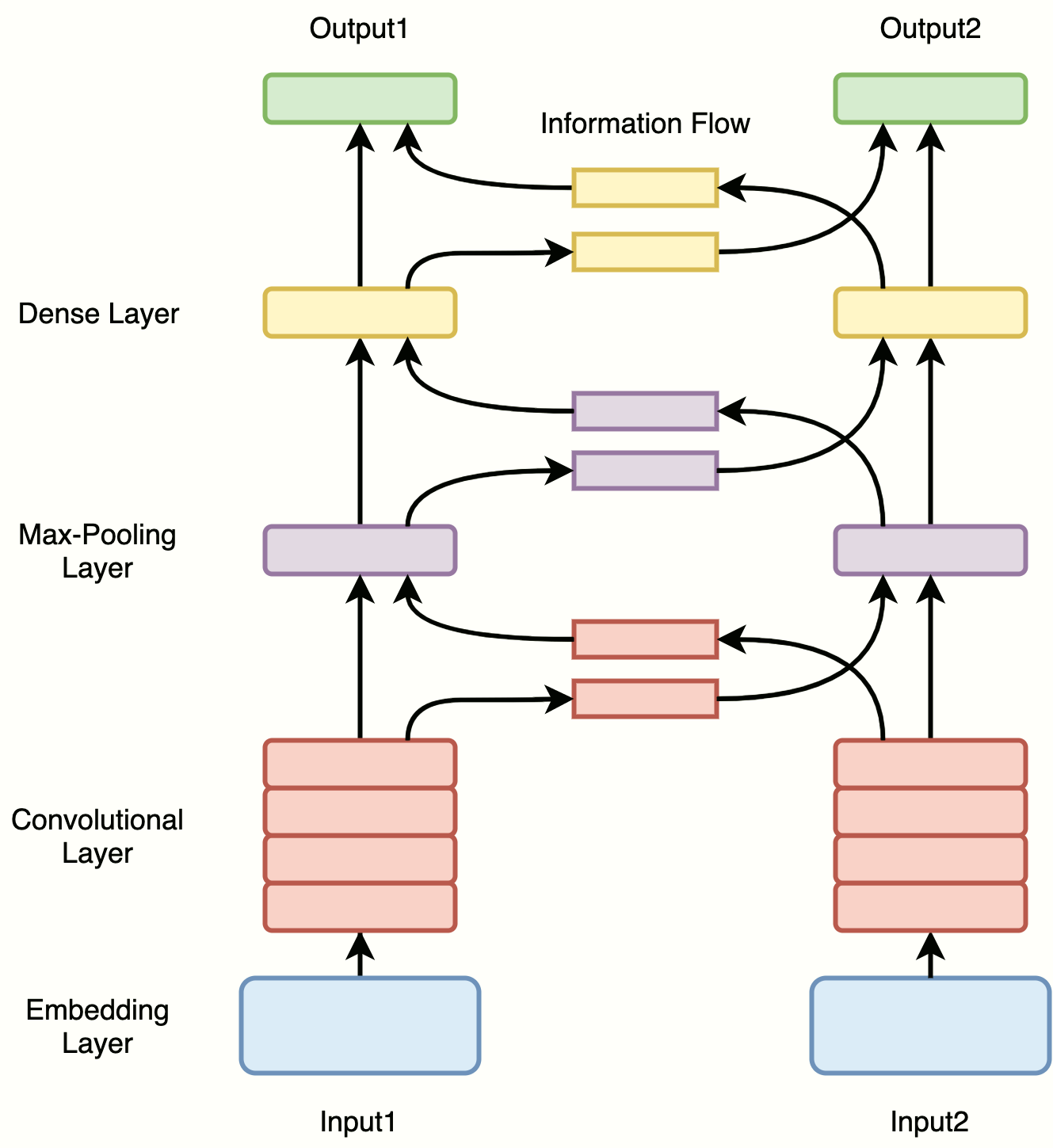} 
  \caption{The structure of the proposed framework.}
  \label{fig:structure}
 \end{figure}

 Same as the works in~\cite{kim2014convolutional}, we apply the convolution operation with the filter $W_{k}\in R^{h\times d}$ to obtain new features in the convolution layer. The operation is depicted in Equation~\ref{equ:convolution}.
 \begin{equation}
 \label{equ:convolution}
 c_{i} = f(W_{k}\cdot X_{i:i+h-1}+b)
 \end{equation}
After that we can obtain a feature map with $\mathbf{c} = [c_{1},c_{2},...,c_{n-h+1}]$.
Then we apply the max-pooling layer to the feature map with $\hat{c} = \max{\mathbf{c}}$ to get the most important feature that is learned with filter $W_{k}$. 
Finally, the dense layer is applied to map the learned features to different classes.
With these three steps, we can get a powerful model in the text classification.

In multitask learning, the gate mechanism can be applied at different levels in order to realize information sharing between different tasks. 
As shown in figure~\ref{fig:structure}, the gates are deployed after the convolutional layer, the max-pooling layer, and the dense layer respectively. In the next subsection, we will give a detailed discussion about gate designing.

\subsection{Information Sharing Gate}

\begin{figure}[htp]
 \centering
   \includegraphics[width=0.94\linewidth]{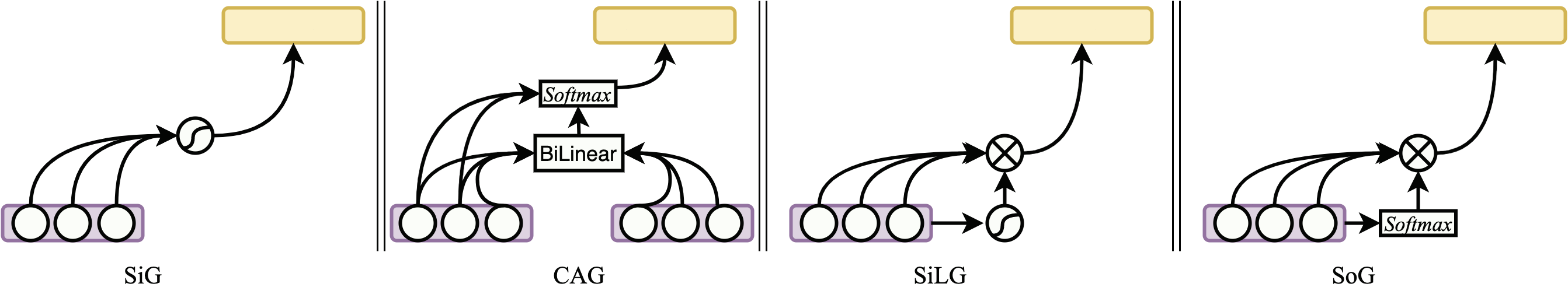} 
  \caption{The sub-net of the information flow unit.}
  \label{fig:if_net}
 \end{figure}

There are different ways to design the gate between two tasks. We need to make sure that the information transfer between the two tasks is useful, but it is difficult to evaluate. 
 To evaluate the information sharing, we apply the cosine similarity between the hidden vectors after the information sharing gate. 
Generally, the similarities between the two tasks should be neither too small nor too big, if the similarity is too small, that means there is no information flowing between two tasks. However, the large similarity denotes all tasks sharing the same structure, which is prone to optimization conflicts. 
In this paper, we discussed three gate mechanisms in controlling the information flow between the tasks of emotion detection and personality trait detection. The structures are shown in Figure~\ref{fig:if_net}.

\textbf{SiG: }The first approach is to pass the information from one network to the other one with Sigmoid Gate (SiG) directly, and it can be shown in equation~\ref{equ:gate_sigmoid}.
 \begin{equation}
 \label{equ:gate_sigmoid}
 \begin{aligned}
& m^{t1}_{i} = \sigma(c^{t1}_{i})\\
&h^{t2}_{i} = c^{t2}_{i} + m^{t1}_{i}\\
 \end{aligned}
 \end{equation}
 Where $\sigma$ denotes the Sigmoid function on hidden value, which works as the gate between the two tasks, while $t1, t2$ are the task indexes. 
 This gate is simple, and it lets useful and useless information pass indiscriminately between the two tasks. Therefore, it is prone to optimization conflicts, and empirical results have validated this phenomenon.

\textbf{CAG: }Another one is the across attention gate (CAG), which is proposed in~\cite{li2020multi}.
CAG considers the contextual information as the extra features in both tasks.
The procedures are described in Equation~\ref{equ:atten_gate}.

\begin{equation}
\begin{aligned}
\label{equ:atten_gate}
& m_{i,j} = c^{t1}_{i}W_{c}c^{t2}_{j}\\
& \alpha_{i,j} = \frac{\exp{m_{i,j}}}{\sum_{k}\exp{m_{i,k}}} \\
& h^{t2}_{i} = c^{t2}_{i} + (\sum_{j}\alpha_{i,j}c^{t2}_{j})\\
\end{aligned}
\end{equation}
where $W_{c}$ denotes the model parameter. 
CAG has considered the features in both tasks in the cross attention calculation, and it should benefit from this cross calculation.
However, the features of each task change dynamically during the optimization process, it will be difficult to get the appropriate optimal feature flow between two tasks with CAG. Also, it takes more time in the CAG calculation.

\textbf{SiLG: } One another gate we discussed here is the Sigmoid weighted linear gate (SiLG), which is proposed in~\cite{elfwing2018sigmoid}, the procedures are shown in Equation~\ref{equ:cl}. This gate shows its effectiveness in reinforcement learning as an approximation of ReLU.

\begin{equation}
\label{equ:cl}
\begin{aligned}
&m^{t1}_{i} = \sigma(c^{t1}_{i})\cdot c^{t1}_{i}\\
&h^{t2}_{i} =c^{t2}_{i} + m^{t1}_{i}\\
\end{aligned}
\end{equation}
While, the disadvantage of SiLG is similar with 
that in SiG, and it cannot select the useful information from a network by passing all information indiscriminately.

\textbf{SoG: } To overcome those problems, we changed the Sigmoid function in SiG to Softmax to construct a selection gate named softmax weight gate SoG, with the result shown in the Equation~\ref{equ:sl}. 
\begin{equation}
\label{equ:sl}
\begin{aligned}
&m^{t1}_{i} = softmax(c^{t1}_{i})\cdot c^{t1}_{i}\\
&h^{t2}_{i} =c^{t2}_{i} + m^{t1}_{i}\\
\end{aligned}
\end{equation}
With this change, the proposed gate mechanism is more effective: 
Compared with CAG, SoG has higher time efficiency in the calculation, compared with SiG and SiLG, SoG can better select features with softmax operation.

\subsection{Meta Multitask Training}
It is the multilabel prediction task for the personality trait detection, therefore, we define the objective function with the multilabel soft margin loss which is described in Equation~\ref{equ:mlsml}
\begin{equation}
\label{equ:mlsml}
\begin{aligned}
L_{Personality} &= -\frac{1}{C}\sum_{i}^{C} y_{i} \cdot \log ((1+\exp(-\hat{y}_{i}))^{-1})\\ 
&+ (1-y_{i})\log (\frac{\exp(-\hat{y}_{i})}{1+\exp(-\hat{y}_{i})})\\
\end{aligned}
\end{equation}
where $C$ is the class number, and it is 5 in our paper. 
 
 However, it is the multiclass prediction task for the emotion detection, and we apply the cross entropy as the objective function which is described in Equation~\ref{equ:xel}.
 \begin{equation}
 \label{equ:xel}
 L_{Emotion} = -\sum_{i}^{C} y_{i} \log (\hat{y}_{i})
 \end{equation}
To optimize the network in a unified framework, we simply add the two loss functions together as the joint function that shows in Equation~\ref{equ:obj},
\begin{equation}
\label{equ:obj}
L_{Multi} = L_{Personality} + L_{Emotion}
\end{equation}
There are different ways to train the model. Generally, each task has its own dataset, therefore, it is necessary to design an adaptive training framework for different subtasks. 
In this paper, like the $k$ shot in MAML~\cite{finn2017model}, we also select the pseudo $k$ shot (i.e., $k$ batches shot) during training, which indicates select $k$ more batches in one task to form the training data pair. Then we update the parameters in a MAML like way. And the algorithm we designed is shown in Algorithm~\ref{alg:1}.

\begin{algorithm}[H] 
 \caption{The optimization method about MAML like in the multitask training.}
 \label{alg:1}
 \begin{algorithmic}[1]
\REQUIRE Personality dataset $D^{p}$ and Emotion dataset $D^{e}$
\REQUIRE Model parameters $\theta = [\theta_{1},\theta_{2},...,\theta_{n}]$, and the learning rate $r$
 \STATE Initialization the model parameters with random values 
 \FOR{select a batch from $D^{e}$}
 \STATE Create a list $\Theta$ and $\mathscr{L}$ for parameter update in each step 
 \STATE Sample $k$ batches $D^{p}_{k}$ from $D_{p}$ 
  \FOR{select a batch from $D_{k}^{p}$}
  		\STATE Obtain the loss $L_{Multi}$ and its gradient descent $\nabla L_{Multi}(f_{\theta})$ 
    \STATE Obtain the updated parameter with $\hat{\theta} = \theta-r\cdot \nabla L_{Multi}(f(\theta))$ and update it to $\theta$ 
    \STATE Stack $\hat{\theta}$ to $\Theta$, and $L_{Multi}$ to $\mathscr{L}$
 \ENDFOR
 \STATE Calculate the gradient to obtain $\nabla \mathscr{L}(\theta)$ and $\nabla \Theta(\theta)$ 
 \STATE Update the parameter $\theta$ with $\nabla\mathscr{L}(\theta) + \nabla \Theta(\theta)$
  \ENDFOR
 \end{algorithmic}
\end{algorithm}

\section{Experiments}
\label{sec:exp}
\subsection{Experiment Settings}

The dataset we applied includes ISEAR~\cite{scherer1994evidence}, TEC~\cite{mohammad2012emotional}, and Personality~\cite{kosinski2013private}. There are 7,666 labeled sentences in ISEAR dataset contains 7 emotions, i.e., \textit{anger, disgust, fear, joy, sadness, shame} and \textit{guilt}. TEC includes 21,051 labeled sentences that are selected in tweets by pre-specified hashtags, i.e., \textit{joy, anger, disgust, surprise, fear} and \textit{sadness}. Personality dataset contains 9,917 multilabeled sentences in the view of the Big Five personality traits. The details about the datasets are listed in Table~\ref{tab:datasets}.
\begin{table*}[htp]
\centering
\begin{tabular}{l|c|c|c}
\hline
Dataset & Total Number & Label Type & Label\\\hline
 ISEAR   &  7,666     & Single & \{anger, disgust, fear, joy, sadness, shame, guilt\} \\\hline
 TEC    &  21,051     & Single & \{joy, anger, disgust, surprise, fear, sadness\} \\ \hline
 Personality & 9,917  & Multiple & \makecell[c]{\{openness, conscientiousness, extraversion, \\agreeableness, and neuroticism\}} \\ \hline
\end{tabular}
\caption{The details of the datasets this paper involved.}
\label{tab:datasets}
\end{table*}
In the experiments, all of the dataset are split randomly in a ratio of 8:2, of which 80\% are training data, and the rest are the test data. 
The embedding dimension is set to 300. 
All the experiments are run on CoLab\footnote{https://colab.research.google.com/} with GPU support, and we run them five times to get the average results in the records.
As we have mentioned, it is the multilabel prediction task for the personality trait detection, and the output of this task is a five-dimension vector after the Sigmoid function, with each value representing a trait in the Five-Factor Model. Therefore, when the value is greater than the threshold (i.e., 0.5), we then regard it as having such a trait. 
While, it is the multiclass prediction task for personality trait detection, and the output of the emotion detection task is a $m$-dimension vector (where $m$ is the emotion categories) after the Softmax function. Therefore, we select the index with the highest value in the $m$-dimension vector as the predicted label. 
To make a comprehensive evaluation of these two tasks, we adopt Accuracy, Precision, Recall, and F1 measurement in the personality trait detection task, and Accuracy in the emotion detection task, respectively.

\subsection{Multitask Learning}
Before the discussion, we list the baseline of the single task in Table~\ref{tab:exp1}. In the baseline, we include the methods like BERT~\cite{devlin2018bert}, CNN~\cite{kim2014convolutional}, LSTM~\cite{hochreiter1997long} those three classical models.

\begin{table*}[htp]
\centering
\begin{tabular}{l|l|l|l|l|l|l|l} \hline
\textbf{}              & \multicolumn{5}{c|}{\textbf{Personality}}      & \textbf{ISEAR}     &\textbf{TEC}        \\ \hline
\textbf{Models} & \textbf{Accuracy} & \textbf{Precision} & \textbf{Recall} & \textbf{F1} &  \textbf{Average}  & \textbf{Accuracy} & \textbf{Accuracy}\\ \hline
BERT Single   & 60.79\%   &   61.53\%  &  77.60\%  & 68.39\% & 67.08\% &  59.28\% &  54.98\%   \\ \hline
 LSTM Single & 61.84\%   & \textbf{64.06\%}  & 75.10\%  & 69.13\% & 67.53\%  & 54.96\%  &55.51\%     \\ \hline
 CNN Single & \textbf{62.23\%}   &  63.17\%  &\textbf{82.60\%} & \textbf{71.54\%} & \textbf{69.89\%}  & \textbf{59.60\%}  &\textbf{56.59\%}    \\ \hline
\end{tabular}
\caption{The baselines results over the datasets we adopted.}
\label{tab:exp1}
\end{table*}
From Table~\ref{tab:exp1}, we can see that CNN Single achieved the best performance in almost all datasets with different evaluation measures. 
That means in the personality trait detection and emotion detection, CNN has a good ability in the feature extraction. And this is also the reason why we choose CNN as the prototype in the design of the multitask model.

Then we carry out experiments to compare the advantages and disadvantages of different variants and models. 
Firstly, to validate the effectiveness of MAML, we make comparisons with the classical optimization method Adam~\cite{kingma2014adam}. 
In the first column of the Table~\ref{tab:exp1_isear} and~\ref{tab:exp1_tec}, the name with ``maml'' indicates that the model is optimized by maml; otherwise, the model is optimized with Adam directly.
In the experiments, the baselines we compared include:
\begin{itemize}
\item \textbf{SiGMTL}: It applies the Sigmoid function as the gate between two different models.
\item \textbf{CAGMTL}: It applies the cross-entropy model as the gate between two different models.
\item \textbf{SiLMTL}: It applies the Sigmoid weighted linear model as the gate between two different models. And it can be treated as that proposed in work~\cite{xiao2018gated}.
\item \textbf{SoGMTL}: It is the multitask learning model we proposed.
\end{itemize}

We then conduct experiments to make comparisons with the different variants and models. The experiment results on dataset ISEAR are shown in Table~\ref{tab:exp1_isear}, and the experiment results on dataset TEC are shown in Table~\ref{tab:exp1_tec}.

\begin{table*}[htp]
\centering
\begin{tabular}{l|l|l|l|l|l|l} \hline
\textbf{}              & \multicolumn{5}{c|}{\textbf{Personality}}        & \textbf{ISEAR}        \\ \hline
\textbf{Models} & \textbf{Accuracy} & \textbf{Precision} & \textbf{Recall} & \textbf{F1} & \textbf{Average}  & \textbf{Accuracy} \\ \hline
 SiGMTL & 61.29\%   &  63.01\%  &71.58\% & 67.61\% &  65.87\%   & 57.20\%   \\ \hline
 SiGMTL+maml & 61.70\%   &  63.62\%  &74.01\% & 67.45\% &66.70\%    & 58.17\%    \\ \hline
CAGMTL   & 62.18\%   &   64.71\%  &  73.39\%  & 69.08\% &67.47\% &  58.13\%      \\ \hline
CAGMTL+maml & 62.41\%   & \textbf{64.96\%}  & 75.27\%  & 68.00\%  & 67.66\%  & 58.55\%      \\ \hline 
SiLGMTL & 60.49\%   &  63.49\%  &68.79\% & 65.96\%   & 64.68\%   & 55.98\%   \\ \hline
SiLGMTL+maml & 60.81\%   &  63.46\%  &70.34\% & 66.15\%  & 65.19\%   & 54.21\%   \\ \hline
SoGMTL  &  62.57\%   & 64.05\%   & 79.96\%   &    71.09\%  & 69.42\%     &   59.03\%    \\ \hline  
SoGMTL+maml  &\textbf{62.77\%}  & 64.22\% &\textbf{83.12\%}  & \textbf{72.45\%} & \textbf{70.64\% }  &  \textbf{59.71\%}   \\ \hline  
\end{tabular}
\caption{The experiment results over the Personality+ISEAR datasets.}
\label{tab:exp1_isear}
\end{table*}

\begin{table*}[htp]
\centering
\begin{tabular}{l|l|l|l|l|l|l} \hline
\textbf{}              & \multicolumn{5}{c|}{\textbf{Personality}}        & \textbf{TEC}        \\ \hline
\textbf{Models} & \textbf{Accuracy} & \textbf{Precision} & \textbf{Recall} & \textbf{F1} &  \textbf{Average}  &  \textbf{Accuracy}         \\ \hline 
 SiGMTL  & 62.24\%   &  64.59\%  &72.83\% & 68.74\%  &67.10\%  & 55.45\%     \\ \hline
SiGMTL+maml & 61.87\%   & 64.76\%  & 72.79\%  & 68.43\% & 66.96\%  & 55.38\%   \\ \hline
CAGMTL  &  62.22\%   & 64.23\%   & 75.81\%   &    69.51\% &67.94\%  &   56.49\%        \\ \hline  
CAGMTL+maml  & 62.86\%  & 64.45\% & 79.29\%  &  71.10\%  &69.43\%  &  57.64\%    \\ \hline  
SiLGMTL  & 61.54\%   &  63.68\%  &72.19\% & 67.62\% & 66.26\%     & 55.32\%    \\ \hline
SiLGMTL+maml & 61.21\%   & 62.26\%  & 77.74\%  & 69.12\%  &67.58\%  & 55.04\%      \\ \hline
SoGMTL  &  62.65\%   & 63.95\%   & 79.40\%   &    70.82\% &69.21\%  &   56.29\%         \\ \hline  
SoGMTL+maml  &\textbf{63.08\% } & \textbf{64.79\%} & \textbf{86.11\%}  &  \textbf{73.93\%}   &\textbf{71.98\%} &  \textbf{57.15\%} \\ \hline  
\end{tabular}
\caption{The experiment results over the Personality+TEC datasets.}
\label{tab:exp1_tec}
\end{table*}

As can be seen from the two tables, SoGMTL+ MAML achieves the best results in personality trait detection under almost all conditions, and the average improvement is about 0.75\% for training with the ISEAR dataset and about 2.09\% for training with the TEC dataset compared with the single task learning with CNN. And it is the same in emotion detection, on the ISEAR dataset, our model improves by 1.32\% compared to single-task learning, and on the TEC dataset, our model improves by 0.56\% compared to single-task learning. 
Especially in the measurement of recall, both datasets showed significant improvement. Compared with the single-task learning using CNN, ISEAR improved by 0.52\% and TEC improved by 3.51\%. 
From these observations, we can conclude that our SoGMTL improves the performance of both tasks in multitask learning through the designed gated mechanism.
Compared to the baselines of the different variants and models in the MTL, we can see that our design achieves the best results in almost all cases, except for the precise measurement of personality trait detection trained with the ISEAR dataset. 
When we make a horizontal comparison between TEC and ISEAR, we find that the performance on TEC dataset is generally greater than that on the ISEAR. 
One reason may be that TEC has three times as much data as ISEAR.  
 In terms of optimization methods, we can see that the MAML method is more effective than the normal method, improving in almost all cases.
 
\begin{figure}[htp]
  \begin{center}
   \includegraphics[width=0.55\linewidth]{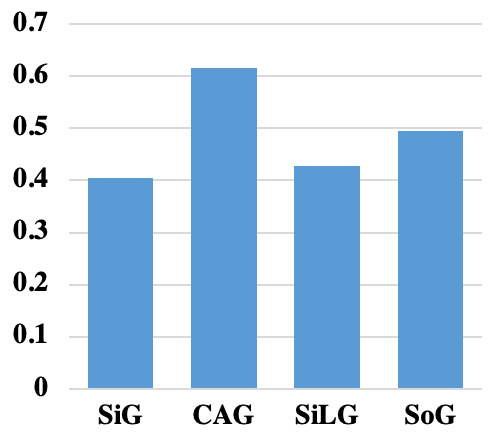}
  \end{center}
  \caption{The running time with different gates, y axis denotes seconds.}
  \label{fig:time}
\end{figure}

 To validate the computation efficiency with these three gate methods, the running time for one epoch is recorded in Figure~\ref{fig:time}.
It can be seen from the figure that SoG performs better than CAG and is competitive with SiG and SiLG. 
Therefore, we can conclude that SoG is the most appropriate approach for personality trait detection and emotion detection in multitask learning.

\subsection{Ablation Study}
To validate the features that learned by the SoG in different levels (i.e., convolutional layer (Conv for short), max-pooling layer (Max-Pooling) and dense layer (Dense), etc.), we only use one SoG gate in the model, and the results are shown in Table~\ref{tab:abl}. 

\begin{table*}[htp]
\centering
\begin{tabular}{l|l|l|l|l|l|l}
\hline
\textbf{}              & \multicolumn{5}{c|}{\textbf{Personality}}        & \textbf{TEC}        \\ \hline
\textbf{Settings} & \textbf{Accuracy} & \textbf{Precision} & \textbf{Recall} & \textbf{F1} & \textbf{Average} & \textbf{Accuracy} \\\hline
Dense     & 62.45\%     & 63.94\%     & 79.30\%     & 70.80\%   & 69.12\%     & 56.55\%    \\\hline
Max-Pooling & 62.45\%      & 63.54\%       & 80.82\%     & 71.13\%   & 69.48\%     & 55.81\%     \\\hline 
Conv     & 62.90\%  & 64.86\%       & 81.05\%     & 72.03\%    & 70.21\%      & 56.95\%     \\\hline
\end{tabular}
\caption{The performance of SoG in different levels that trained with TEC dataset.}
\label{tab:abl}
\end{table*}
It can be seen from the table that the performance of SoG in the convolutional layer is the best, followed by the performance in the max-pooling layer. 
That means that the SoG in the convolutional layer is the most important in the information flow control. 
Furthermore, to quantify the degree of SoG's control over information flows at different locations, we calculate the cosine similarity between the latent vectors of two tasks in the convolutional layer (i.e., \textbf{Sim1}) and dense layer (i.e., \textbf{Sim2}) respectively. And the results are shown in Table~\ref{tab:sim}.

\begin{table}[H]
\centering
\begin{tabular}{l|c|c}
\hline
\textbf{Variants} & \textbf{Sim1} & \textbf{Sim2} \\\hline
Dense    & 0.21 (-0.13) & 0.47 (-0.05)\\\hline
Max-Pooling & 0.14 (-0.20) & 0.33 (-0.19)\\\hline
Conv    & 0.32 (-0.02) & 0.47 (-0.05) \\\hline
SoGMTL   & 0.34 & 0.52 \\\hline
\end{tabular}
\caption{The cosine similarity between the latent vector of the two tasks in the convolutioanl layer and dense layer respectively when SoG is placed at different levels.}
\label{tab:sim}
\end{table}
SoGMTL is assumed to be the "best" status of information control in multitask learning.
Therefore, it can be concluded from the table that when SoG is only placed in the dense layer, the difference between Sim1 and the best is -0.13, and the difference between Sim2 and the best is -0.05.
However, both gaps become extremely small when SoG is only placed in the convolutional layer. 	
Thus, we can know that more useful information is shared via the SoG in the convolutional layer, followed by the max-pooling layer, and less of them are shared with the dense layer. 
That is why the average performance is the best when SoG is only placed in the convolutional layer shown in Table~\ref{tab:abl}.

\subsection{Case Study}
To make a clear understanding of SoGMTL's performance in multitask learning, a case study is conducted, and the results are shown in Table~\ref{tab:cases}.
 
\begin{table*}[htp]
\centering
\begin{tabular}{c|l|l|c|c}
\hline
     Case    & Category      & \makecell[c]{ Sentence} & True Label & Predicted \\\hline
{\multirow{2}{*}{1}} & Personality &  \makecell[l]{ Is still awake at 3:30 .\\ oh me.}   & Neuroticism, Openness     &   \begin{minipage}{.2\textwidth} \includegraphics[width=34mm, height=17mm]{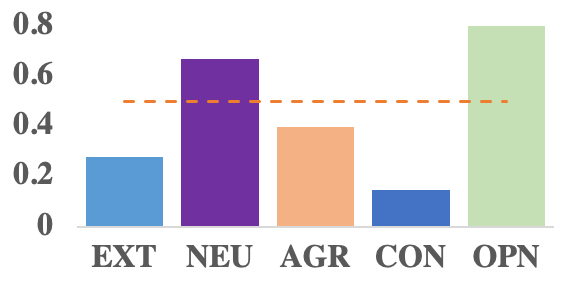}  \end{minipage}  \\\cline{2-5}
        & Emotion   &  \makecell[l]{ I'm home watchin this sad\\ movie. Missing college.}  &  Sadness    &   \begin{minipage}{.2\textwidth} \includegraphics[width=34mm, height=17mm]{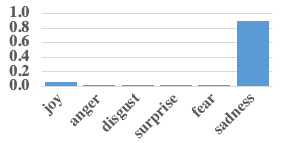}  \end{minipage}    \\\hline
\multirow{2}{*}{2}           & Personality &  \makecell[l]{Damn you's a sexy bitch \\DAMN GIRL!!! }   &     \makecell[c]{ Neuroticism, Agreeableness, \\Openness }   &\begin{minipage}{.2\textwidth} \includegraphics[width=34mm, height=17mm]{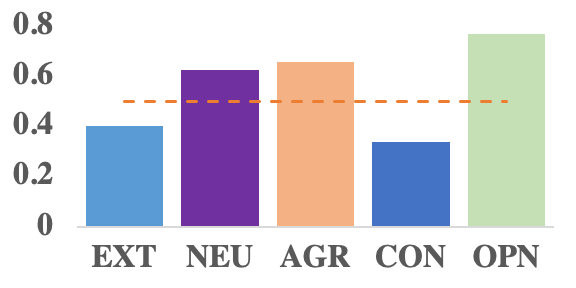}  \end{minipage}      \\\cline{2-5}
                    & Emotion   &   \makecell[l]{ Holding my fucking \\tongue}	 &  Anger    & \begin{minipage}{.2\textwidth} \includegraphics[width=34mm, height=17mm]{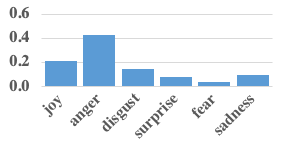}  \end{minipage}     \\\hline
                    
\end{tabular}
\caption{The case study about the SoGMTL's performance in the personality trait detection and emotion detection. The dashed line in the predicted image indicates the threshold in the personality trait detection, it is 0.5 in our paper.}
\label{tab:cases}
\end{table*}
 
Personality trait detection is a multilabel classification problem, so there are multiple labels in a detection. When the value is greater than a threshold (i.e., 0.5), it indicates the personality has this trait.
And the cases in the Table~\ref{tab:cases} show the effectiveness of SoGMTL in this task by having clear boundaries in the prediction. 
However, emotion detection is a multiclass classification problem~\cite{camnt6}. In the prediction, the label is determined by the maximum value. As can be seen from the image in Table~\ref{tab:cases}, there is an obvious difference between the maximum value and other values.
In other words, our proposed model can be well applied to personality trait detection and emotion detection in a multitask learning approach.

\section{Conclusion}
\label{sec:con}
In this work, we designed a multitask learning framework in personality trait detection and emotion detection. Also, we have discussed and validated the effective sharing way between two tasks. Finally, we designed a MAML-like training method when there is more than one data source. Empirical results show the effectiveness of our proposed framework and optimization method. In future work, more contextual information like personal preference, current location, etc., will be considered. 

\section*{Acknowledgments}
The research undertaken in the Continental-NTU Corporate Lab is supported by the Singapore Government.

\bibliography{references}
\bibliographystyle{plain}
\end{document}